\documentclass{article}
\usepackage[utf8]{inputenc}
\usepackage{main}
\usepackage{microtype}
\usepackage{graphicx}
\usepackage{times}
\usepackage{latexsym}
\usepackage{amsmath}
\usepackage{float}
\usepackage{footnote}
\usepackage{enumitem}
\usepackage{bm}
\usepackage{arydshln}
\usepackage{booktabs}
\usepackage{multicol}
\usepackage{multirow}
\usepackage{color}
\usepackage{xcolor}     
\usepackage{colortbl}
\usepackage{bbding}
\usepackage{makecell}
\usepackage{mathtools}
\usepackage{imakeidx}
\usepackage{longtable}
\usepackage{wrapfig}
\makeindex
\usepackage{arydshln}
\usepackage{lipsum}
\usepackage{natbib}
\usepackage[toc]{multitoc}
\usepackage[edges]{forest}
\usepackage[normalem]{ulem}
\definecolor{mydarkblue}{rgb}{0,0.08,0.45}
\usepackage[colorlinks=true,linkcolor=mydarkblue,citecolor=mydarkblue,filecolor=mydarkblue,urlcolor=mydarkblue]{hyperref}
\usepackage{caption}
\usepackage{CJKutf8}
\usepackage{awesomebox} % for infobox
\usepackage{bbding}
\usepackage[most]{tcolorbox}
\usepackage{graphicx}
\usepackage{amsmath}
\usepackage{booktabs}
\usepackage{multirow}
\usepackage{pifont}
\usepackage{caption}
\usepackage{subcaption}

\definecolor{GreenCheck}{RGB}{0, 102, 51}
\definecolor{darkGreen}{rgb}{0.2,0.5,0.2}
\definecolor{dodgerBlue}{rgb}{0.19, 0.55, 0.91}
\definecolor{darkRed}{rgb}{1.0, 0.03, 0.0}
\definecolor{weak}{RGB}{208,100,100}
\definecolor{icl}{RGB}{95,148,207}
\definecolor{hybrid}{RGB}{140,185,110}
\definecolor{strongceiling}{RGB}{150,150,150}
\newcommand{\y}{\textcolor{GreenCheck}{\ding{52}}}
\newcommand{\n}{\textcolor{red}{\ding{56}}}
\newcommand{\notmust}{\textcolor{gray}{\textbf{--}}}
\newcommand{\M}{\mathcal{M}}
\newcommand{\Q}{\mathcal{Q}}
\newcommand{\D}{\mathcal{D}}

\usepackage[tikz]{bclogo}
\usepackage[framemethod=tikz]{mdframed}
\definecolor{bgblue}{RGB}{245,243,253}
\definecolor{ttblue}{RGB}{91,194,224}

\mdfdefinestyle{mystyle}{%
  rightline=true,
  innerleftmargin=10,
  innerrightmargin=10,
  outerlinewidth=3pt,
  topline=false,
  rightline=true,
  bottomline=false,
  skipabove=\topsep,
  skipbelow=\topsep
}

\newtcolorbox{myboxi}[1][]{
  breakable,
  title=#1,
  colback=red!5,
  colbacktitle=red!5,
  coltitle=black,
  fonttitle=\bfseries,
  bottomrule=0pt,
  toprule=0pt,
  leftrule=2pt,
  rightrule=2pt,
  titlerule=0pt,
  arc=0pt,
  outer arc=0pt,
  colframe=red,
}

\newtcolorbox{myboxnote}[1][]{
  breakable,
  title=#1,
  colback=orange!0,
  colbacktitle=orange!0,
  coltitle=black,
  fonttitle=\bfseries,
  bottomrule=0pt,
  toprule=0pt,
  leftrule=2pt,
  rightrule=2pt,
  titlerule=0pt,
  arc=0pt,
  outer arc=0pt,
  colframe=orange,
}

\newtcolorbox{myboxii}[1][]{
  breakable,
  freelance,
  title=#1,
  colback=white,
  colbacktitle=white,
  coltitle=black,
  fonttitle=\bfseries,
  bottomrule=0pt,
  boxrule=0pt,
  colframe=white,
  overlay unbroken and first={
  \draw[red!75!black,line width=3pt]
    ([xshift=5pt]frame.north west) -- 
    (frame.north west) -- 
    (frame.south west);
  \draw[red!75!black,line width=3pt]
    ([xshift=-5pt]frame.north east) -- 
    (frame.north east) -- 
    (frame.south east);
  },
  overlay unbroken app={
  \draw[red!75!black,line width=3pt,line cap=rect]
    (frame.south west) -- 
    ([xshift=5pt]frame.south west);
  \draw[red!75!black,line width=3pt,line cap=rect]
    (frame.south east) -- 
    ([xshift=-5pt]frame.south east);
  },
  overlay middle and last={
  \draw[red!75!black,line width=3pt]
    (frame.north west) -- 
    (frame.south west);
  \draw[red!75!black,line width=3pt]
    (frame.north east) -- 
    (frame.south east);
  },
  overlay last app={
  \draw[red!75!black,line width=3pt,line cap=rect]
    (frame.south west) --
    ([xshift=5pt]frame.south west);
  \draw[red!75!black,line width=3pt,line cap=rect]
    (frame.south east) --
    ([xshift=-5pt]frame.south east);
  },
}

\usepackage{fancyhdr} % to change header and footers
\usepackage{blindtext} % to quickly get a full document

\DeclareCaptionFont{black}{\color{black}}

\definecolor{myblue}{rgb}{0.9, 0.1, 0.94}
\definecolor{mygreen}{rgb}{0.64, 0.56, 0.88}
\definecolor{myyellow}{rgb}{0.68, 0.6, 0.1}
\definecolor{fancygreen}{rgb}{0.33, 0.68, 0.20}
\definecolor{salmon}{rgb}{0.94, 0.52, 0.49}
\definecolor{tablegreen}{rgb}{0.82, 0.94, 0.75}
\definecolor{tableblue}{rgb}{0.81, 0.90, 0.94}
\definecolor{tablered}{rgb}{0.97, 0.85, 0.85}
\definecolor{tableorange}{rgb}{0.96, 0.85, 0.81}

\newenvironment{itemize*}%
 {\leftmargini=10pt\begin{itemize}%
  \setlength{\itemsep}{0pt}%
  \setlength{\parskip}{0pt}%
  }%
 {\end{itemize}}
\newenvironment{enumerate*}%
 {\begin{enumerate}%
  \setlength{\itemsep}{0pt}%
  \setlength{\parskip}{0pt}}%
 {\end{enumerate}}

\usepackage{xcolor}
\usepackage{listings}

\newcommand\JSONnumbervaluestyle{\color{blue}}
\newcommand\JSONstringvaluestyle{\color{red}}

\newif\ifcolonfoundonthisline

\makeatletter

\lstdefinestyle{json}
{
  showstringspaces    = false,
  keywords            = {false,true},
  alsoletter          = 0123456789.,
  morestring          = [s]{"}{"},
  stringstyle         = \ifcolonfoundonthisline\JSONstringvaluestyle\fi,
  MoreSelectCharTable =%
    \lst@DefSaveDef{`:}\colon@json{\processColon@json},
  basicstyle          = \ttfamily,
  keywordstyle        = \ttfamily\bfseries,
}

\newcommand\processColon@json{%
  \colon@json%
  \ifnum\lst@mode=\lst@Pmode%
    \global\colonfoundonthislinetrue%
  \fi
}

\lst@AddToHook{Output}{%
  \ifcolonfoundonthisline%
    \ifnum\lst@mode=\lst@Pmode%
      \def\lst@thestyle{\JSONnumbervaluestyle}%
    \fi
  \fi
  \lsthk@DetectKeywords% 
}

\lst@AddToHook{EOL}%
  {\global\colonfoundonthislinefalse}

\makeatother

\usepackage{etoolbox}
\usepackage{natbib}
\usepackage{url}
\newcounter{bibcount}
\makeatletter
\patchcmd{\@lbibitem}{\item[}{\item[\hfil\stepcounter{bibcount}{[\thebibcount]}}{}{}
\renewcommand\NAT@bibsetup%
  [1]{\setlength{\leftmargin}{\bibhang}\setlength{\itemindent}{-\parindent}%
      \setlength{\itemsep}{\bibsep}\setlength{\parsep}{\z@}}
\makeatother

\begin{document}

\title{Weak-to-Strong Reasoning}

\author{
    Yuqing Yang\textsuperscript{\rm{2,4}} \quad Yan Ma\textsuperscript{\rm{2,3,4}} \quad \textbf{Pengfei Liu}\textsuperscript{\rm{1,3,4}}\thanks{Corresponding author.} \\
    \textsuperscript{1}Shanghai Jiao Tong University \
    \textsuperscript{2}Fudan University \\
    \textsuperscript{3}Shanghai AI Laboratory \
    \textsuperscript{4}Generative AI Research Lab (GAIR) \\
    \texttt{\{yuqingyang21, yanma23\}@m.fudan.edu.cn} \quad \texttt{pengfei@sjtu.edu.cn}
}

\maketitle
% \footnotetext[1]{\,Corresponding Author.}
\thispagestyle{fancy}
\fancyhead{}
\lhead{\includegraphics[height=0.67cm]{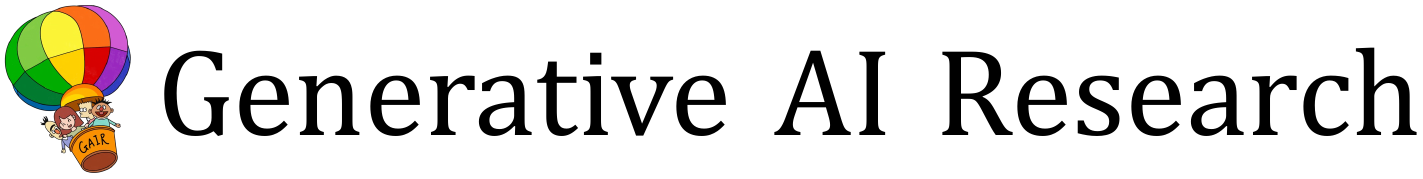}}
\renewcommand{\headrulewidth}{0pt}
\setlength{\headsep}{0mm}

\begin{abstract}

When large language models (LLMs) exceed human-level capabilities, it becomes increasingly challenging to provide full-scale and accurate supervision for these models. Weak-to-strong learning, which leverages a less capable model to unlock the latent abilities of a stronger model, proves valuable in this context. Yet, the efficacy of this approach for complex reasoning tasks is still untested. Furthermore, tackling reasoning tasks under the weak-to-strong setting currently lacks efficient methods to avoid blindly imitating the weak supervisor including its errors. In this paper, we introduce a progressive learning framework that \textbf{enables the strong model to autonomously refine its training data, without requiring input from either a more advanced model or human-annotated data}. This framework begins with supervised fine-tuning on a selective small but high-quality dataset, followed by preference optimization on contrastive samples identified by the strong model itself. Extensive experiments on the GSM8K and MATH datasets demonstrate that our method significantly enhances the reasoning capabilities of Llama2-70b using three separate weak models. This method is further validated in a forward-looking experimental setup, \textbf{where Llama3-8b-instruct effectively supervises Llama3-70b on the highly challenging OlympicArena dataset}. This work paves the way for a more scalable and sophisticated strategy to enhance AI reasoning powers. All relevant code and resources are available in \url{https://github.com/GAIR-NLP/weak-to-strong-reasoning}.

\end{abstract}

\begin{figure}[ht]
    \vspace{30px}
    \centering
    \begin{subfigure}[b]{0.45\textwidth}
        \centering
        \includegraphics[width=\textwidth]{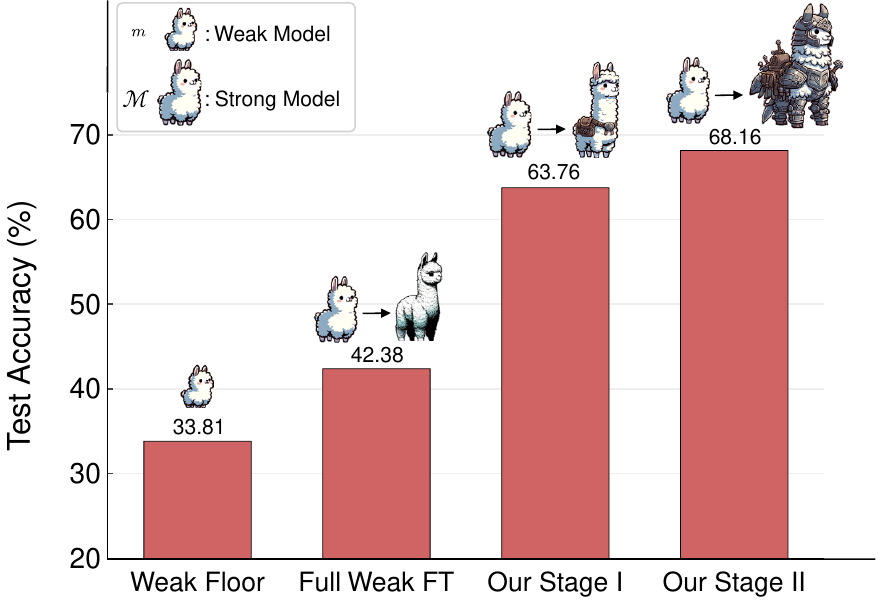}
        \caption{\centering Llama2-7b \raisebox{-.2ex}{\includegraphics[height=2ex]{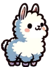}} supervises Llama2-70b \raisebox{-.2ex}{\includegraphics[height=3ex]{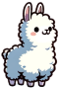}} \newline on GSM8K~\citep{DBLP:journals/corr/abs-2110-14168}.}
        \label{fig:Fig0_gsm8k}
    \end{subfigure}
    \begin{subfigure}[b]{0.45\textwidth}
        \centering
        \includegraphics[width=\textwidth]{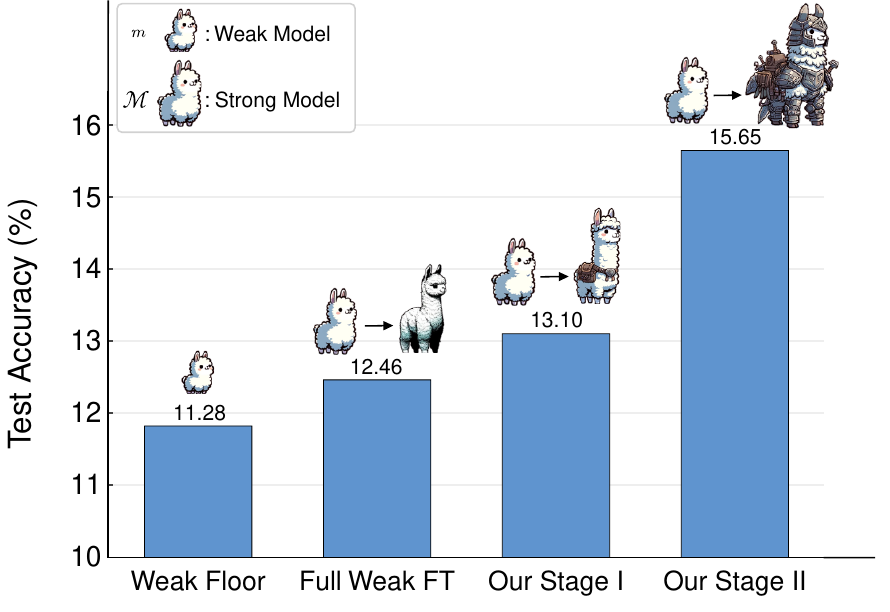}
        \caption{\centering Llama3-8b-instruct \raisebox{-.2ex}{\includegraphics[height=2ex]{figures/llama_m.png}} supervises Llama3-70b \raisebox{-.2ex}{\includegraphics[height=3ex]{figures/llama_bigm.png}} \newline on OlympicArena~\citep{huang2024olympicarenabenchmarkingmultidisciplinecognitive}.}
        \label{fig:Fig0_olympic}
    \end{subfigure}
    \caption{(a): Test accuracy on GSM8K using Llama2-7b to supervise Llama2-70b. (b): Test accuracy on OlympicArena using Llama3-8b-instruct to supervise Llama3-70b. ``Weak Floor'' refers to the results of the weak model. ``Full Weak FT'' refers to the results of the baseline where the strong model is naively fine-tuned on the full dataset generated by the weak model. ``Our Stage I'' represents the results from the first stage of supervised fine-tuning using our proposed weak-to-strong method. Note that our method in Stage I produces three variants of enhanced strong models and we present the best results here. ``Our Stage II'' denotes the results from the second stage of preference optimization using our method.}
    \label{fig:Fig0}
\end{figure}

\newpage

\pagestyle{fancy}
\lhead{Weak-to-Strong Reasoning}
\renewcommand{\headrulewidth}{0.7pt}
\setlength{\headsep}{5mm}

\section{Introduction}

\begin{quote}
``\textit{A student need not be inferior to the teacher; a teacher need not be wiser than the student.}'' \\
— \underline{On Teachers}
\end{quote}

\begin{wrapfigure}{r}{0.5\textwidth}
    \vspace{-3mm}
    \centering
    \begin{minipage}{0.5\textwidth}
    \includegraphics[width=\linewidth]{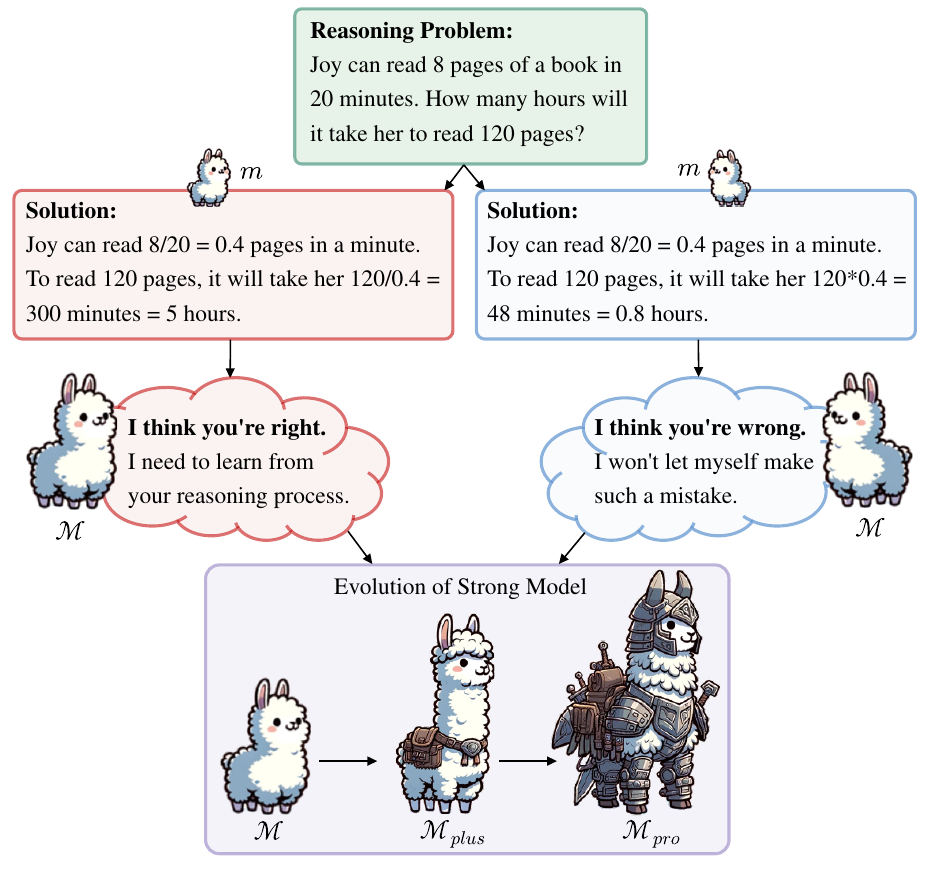}
    \caption{\small Illustration of weak-to-strong reasoning through the strong model self-refining its training data.}
    \label{fig:fig1}
    \end{minipage}
    \vspace{-3mm}
\end{wrapfigure}

As the pursuit of Artificial General Intelligence (AGI) advances, the creation of superintelligent systems—models that exceed human cognitive capabilities—remains a key ambition within the field \citep{Robert2017SuperintelligencePD,openai_governance_2023,DBLP:journals/corr/abs-2403-14683}. This quest introduces a host of challenges, especially concerning the \textit{supervision} and \textit{learning paradigms} for these advanced AI models. Conventional supervision methods, which typically depend on human oversight \citep{DBLP:conf/nips/ChristianoLBMLA17,DBLP:conf/nips/Ouyang0JAWMZASR22,DBLP:journals/corr/abs-2403-09472} or guidance (i.e., distilled knowledge) from more advanced models \citep{DBLP:journals/corr/abs-2212-08073,DBLP:journals/corr/abs-2309-00267,DBLP:journals/corr/abs-2304-03277}, become inadequate as the capabilities of AI exceed those of their supervisors \citep{DBLP:journals/corr/abs-2211-03540,DBLP:journals/corr/abs-2402-00667}. To address this issue, we focus on the \textit{weak-to-strong learning} paradigm \citep{DBLP:journals/corr/abs-2312-09390}, which operates under a unique task setting where only a less capable model and a stronger\footnote{Similar to~\citet{DBLP:journals/corr/abs-2312-09390}, we define ``strong model'' in the context of LLMs, taking into account their characteristics—that is, LLMs often contain the knowledge and capabilities needed to perform specific tasks, but these have not yet been fully elicited~\cite{zhou2024lima}. Typically, it refers to stronger and larger pre-trained language models whose capabilities have not been fully realized yet.} but not fully utilized model are available.

The central question of weak-to-strong learning is whether models with limited capabilities can effectively guide the development of more advanced, stronger models. Previous studies by \citet{DBLP:journals/corr/abs-2312-09390} have demonstrated the feasibility of it in classification, chess, and reward modeling tasks. However, the applicability of this setup to more complex reasoning tasks, which demand more than mere extrapolation or pattern recognition, remains an open question. Complex reasoning represents a key aspect of human cognition, crucial for assessing whether LLMs can emulate or surpass human-like capabilities in comprehending the world, making decisions, and solving problems \citep{DBLP:conf/acl/QiaoO0CYDTHC23,DBLP:conf/acl/0009C23,DBLP:journals/corr/abs-2307-03109}. Given the complexity and the critical nature of these tasks, applying the weak-to-strong learning framework to advanced reasoning challenges is essential, particularly within the broader context of achieving superintelligence.

Although \citet{DBLP:journals/corr/abs-2312-09390} suggest that naively fine-tuning strong models on the full set of noisy data produced by weak models, named \textit{full weak fine-tuning}, can consistently improve their performance over the weaker counterparts, this approach is still far from recovering the full capabilities of strong models, and our experiments show that it loses effectiveness when facing more complex reasoning challenges. They also propose an auxiliary confidence loss to mitigate the issue of strong models imitating the errors of their supervisors. However, this method is tailored to classification tasks with a set of fixed labels and does not naturally extend to open-ended generation tasks including reasoning. Currently, there is a lack of effective methods beyond naive fine-tuning to prevent the overfit of weak errors and to further elicit the intrinsic reasoning abilities of strong models within the \textbf{weak-to-strong reasoning} framework.

To achieve the above goal, we introduce a progressive refinement learning framework, guided by the principle that a model can enhance its capabilities more effectively by initially focusing on smaller, more reliable subsets of data, and then iteratively expanding its learning scope, as illustrated in Fig.~\ref{fig:fig1}. In the first stage, we hypothesize that it is more advantageous to utilize smaller quantities of data that are likely to be more accurate. We achieve this by combining weak data, generated by the less capable model, with data self-generated by the more advanced model through in-context learning. This blend is then used to selectively curate datasets for subsequent supervised fine-tuning. In the second stage, upon having developed a strong model with improved reasoning capabilities, we utilize its ability to construct contrastive samples for preference optimization \citep{DBLP:conf/nips/RafailovSMMEF23,DBLP:journals/corr/abs-2403-07691} and enable the model to learn effectively from the errors of the weaker model. 

In implementation, we employ Llama2-70b \citep{DBLP:journals/corr/abs-2307-09288} as the strong model, test three separate weak models: Llama2-7b, Gemma-2b \citep{DBLP:journals/corr/abs-2403-08295}, and Mistral-7b \citep{DBLP:journals/corr/abs-2310-06825}, and conduct experiments on the commonly used math reasoning datasets GSM8K \citep{DBLP:journals/corr/abs-2110-14168} and MATH \citep{DBLP:conf/nips/HendrycksBKABTS21}. Experimental results reveal that:
\begin{enumerate*}
    \item Full weak fine-tuning, while effective in classification tasks, falls short for complex reasoning tasks.
    \item Our proposed method significantly outperforms full weak fine-tuning method, achieving a \textbf{26.99-point} improvement on GSM8K when supervised solely by the weak model (i.e., Gemma-2b) after the first stage of training ($\M \to \M_{\text{plus}}$), and further enhances performance by an additional \textbf{8.49} points through preference optimization without knowing the gold answer ($\M_\text{plus} \to \M_\text{pro}$).\
    \item Our proposed preference optimization phase enables the strong model to learn from errors made by the weak supervisor, \textbf{ultimately surpassing the strong model fine-tuned on gold-standard solutions} (i.e., strong ceiling) in challenging scenarios, such as level 4-5 MATH problems.
\end{enumerate*}

To more accurately approximate future scenarios, we conduct additional \textbf{experiments on OlympicArena \citep{huang2024olympicarenabenchmarkingmultidisciplinecognitive}, an extremely challenging dataset with \emph{no} definitive ground truth answers}. Llama3-8b-instruct \citep{llama3modelcard}, despite its smaller size, has been aligned and proven to effectively supervise the larger Llama3-70b, whose potential has not yet been fully realized. Moreover, our proposed two-stage training approach outperforms full weak fine-tuning by \textbf{3.19} points.

\section{Preliminaries}

\subsection{Typical Learning Paradigms for LLMs}

\begin{wraptable}{r}{0.50\textwidth}
    \vspace{-3mm}
    \centering
    \small
    \begin{tabular}{lcc}
    \toprule
     & \textbf{G.T. Answer} & \textbf{Stronger Model} \\
    \midrule
    Generic-supervised & \y & \notmust \\
    Distillation-based & \n & \y \\
    Self-improvement & \y & \notmust \\
    Semi-supervised & \y & \notmust \\
    Weak-to-strong & \n & \n \\
    \bottomrule
    \end{tabular}
    \caption{\small Typical Learning Paradigms for LLMs. ``\y'' and ``\n'' indicate whether supervision is required, and ``\notmust'' indicates it is optional. ``G.T.'' represents Ground Truth.}
    \label{tab:learning_paradigm}
    \vspace{-3mm}
\end{wraptable}

We outline common learning paradigms in large model training, primarily characterized by the need for ground truth answers and supervision from stronger models as shown in Tab.~\ref{tab:learning_paradigm}.

\paragraph{Generic-Supervised Learning}
When training LLMs, it is ideal to have a sufficient amount of training data with ground truth answers, which we refer to as \textit{generic-supervised learning} paradigm~\cite{DBLP:conf/nips/Ouyang0JAWMZASR22,DBLP:journals/corr/abs-2308-01825}. However, acquiring such data is often label-intensive and can sometimes be impossible. As a result, various learning paradigms have emerged to reduce the effects of data quality and quantity while still improving performance. 

\paragraph{Distillation-based Learning}
In the current context, to enhance a strong model like Llama2-70b, improvements can still be made by seeking help to a stronger model like GPT-4 \citep{DBLP:journals/corr/abs-2303-08774}, even without ground truth. Hence, many existing works suggest that a stronger model acts as a teacher model to provide specific feedback to improve the targeted model \citep{DBLP:journals/corr/abs-2309-00267,DBLP:journals/corr/abs-2304-03277,DBLP:journals/corr/abs-2310-20689,DBLP:journals/corr/abs-2306-13649,DBLP:journals/corr/abs-2310-10477}. This paradigm can be viewed as distilling the stronger teacher model's knowledge. Nonetheless, merely imitating the teacher model is not a long-term solution; imitation models only slightly close the performance gap to the teacher model on tasks not well-represented in the imitation data \citep{DBLP:journals/corr/abs-2305-15717}. Furthermore, distillation learning primarily benefits models that are less capable than the teacher model.

\paragraph{Self-Improvement Learning}
Considering the high costs of annotating training data by humans or stronger proprietary models, a line of works relies on the \textit{correct} responses generated by the model itself to update it. For example, \citet{DBLP:conf/nips/ZelikmanWMG22,DBLP:journals/corr/abs-2308-01825,DBLP:journals/corr/abs-2312-06585,DBLP:journals/corr/abs-2402-06457} filter solutions according to the correctness of final answers, while \citet{DBLP:journals/corr/abs-2305-20050,DBLP:journals/corr/abs-2404-07965} employ reward models trained on gold annotations to score self-generated content. It is evident that, whether using binary labels or fine-grained feedback, this paradigm still requires ground truth to assess the usability of the model's self-generated responses. Without ground truth answers, self-improvement leads to minimal performance gains and may even degrade performance \citep{DBLP:journals/corr/abs-2310-01798,DBLP:journals/corr/abs-2311-08516}.

\paragraph{Semi-Supervised Learning}
Gaining insights from semi-supervised learning within the domain of traditional machine learning, another type of LLM learning depends not on extensive labeling but instead on a small, high-quality seed dataset. \citet{tong2024optimizing} have demonstrated improvement by learning differences between self-generated responses and expert-annotated responses. We also include the trending research topic of \textit{easy-to-hard generalization} \citep{DBLP:journals/corr/abs-2401-06751,DBLP:journals/corr/abs-2403-09472} in this category, where models are trained to tackle complex tasks by learning from human annotations on easier tasks. This series of research inevitably requires access to a small yet high-quality set of standard answers.

\paragraph{Weak-to-Strong Learning}
In scenarios where models surpass human capabilities, the challenge of providing comprehensive and precise supervision for complex tasks intensifies, particularly as \textbf{no ground truth exists, nor a superior model for supervisory guidance}. This absence underscores the critical importance of \emph{weak-to-strong learning} approaches. Such methods uniquely leverage weaker supervisory signals to recover latent knowledge from already powerful models. For example, fine-tuning GPT-4 with a GPT-2-level supervisor can recover close to GPT-3.5-level performance on certain tasks~\cite{DBLP:journals/corr/abs-2312-09390}. This strategy holds profound implications for advancing human societal progress by equipping LLMs with the capabilities to address currently unsolvable mathematical and physical challenges. Unlike other learning paradigms, weak-to-strong learning operates under comparatively relaxed conditions, opening expansive opportunities for exploration and innovation.

\subsection{Weak-to-Strong Reasoning Setup}

\begin{wraptable}{r}{0.50\textwidth}
    \vspace{-3mm}
    \centering
    \small
    \begin{tabular}{c|ccc}
    \toprule
    \textbf{Role} & weak model & strong model & task question \\
    \midrule
    \multirow{2}{*}{\textbf{Analogue}} & Llama2-7b & \multirow{2}{*}{Llama2-70b} & $\Q \in \text{GSM8K}$ \\
    & {\tiny + SFT($\D_{\text{gold}, 1}$)} & & $\mathrel{\phantom{=}} \in \text{MATH}$ \\
    \bottomrule
    \end{tabular}
    \caption{\small Weak-to-Strong Reasoning Setup.}
    \label{tab:setup}
    % \vspace{-3mm}
\end{wraptable}

In this paper, we address reasoning tasks in the weak-to-strong setting, as illustrated in Tab.~\ref{tab:setup}. First, we examine mathematical reasoning tasks, such as those in GSM8k and MATH. These tasks require each step of the reasoning process to demonstrate fundamental mathematical problem-solving skills, including problem comprehension and algebraic operations, and build upon the previous steps. This imposes higher demands on the model's learning and generalization capabilities. Unlike classification tasks, where models can rely on superficial pattern extrapolation or recognition, reasoning tasks offer minimal benefit from guessing. Then, we use \textbf{a weak model} (e.g., Llama2-7b) with a certain degree of mathematical problem-solving ability,\footnote{Otherwise, the weak model can hardly provide useful supervision.} denoted as $m$. This model acts analogously to human supervisors with limited expertise in the era of superintelligence. Besides, we only have \textbf{a set of questions} $\mathcal{Q} = \{q_i\}$ without ground truth answers and the goal is to improve the reasoning capability of \textbf{a strong model} $\M$ (e.g., Llama2-70b). To implement this, following \citet{DBLP:journals/corr/abs-2312-09390}, we randomly divide the original training set into two equal parts, $\D_{\text{gold}, 1}$ and $\D_{\text{gold}, 2}$. The weak model is initially fine-tuned using $\D_{\text{gold}, 1}$ where the gold solutions are available, resulting in a weak model with some problem-solving capability, i.e. $m$. In contrast, the strong model can only access the questions from $\D_{\text{gold}, 2}$, without reasoning chains or final answers, i.e., $\mathcal{Q}$.

\section{Methodology}

\begin{figure*}[t] \centering
    \vspace{-3mm}
    \includegraphics[width=\linewidth]{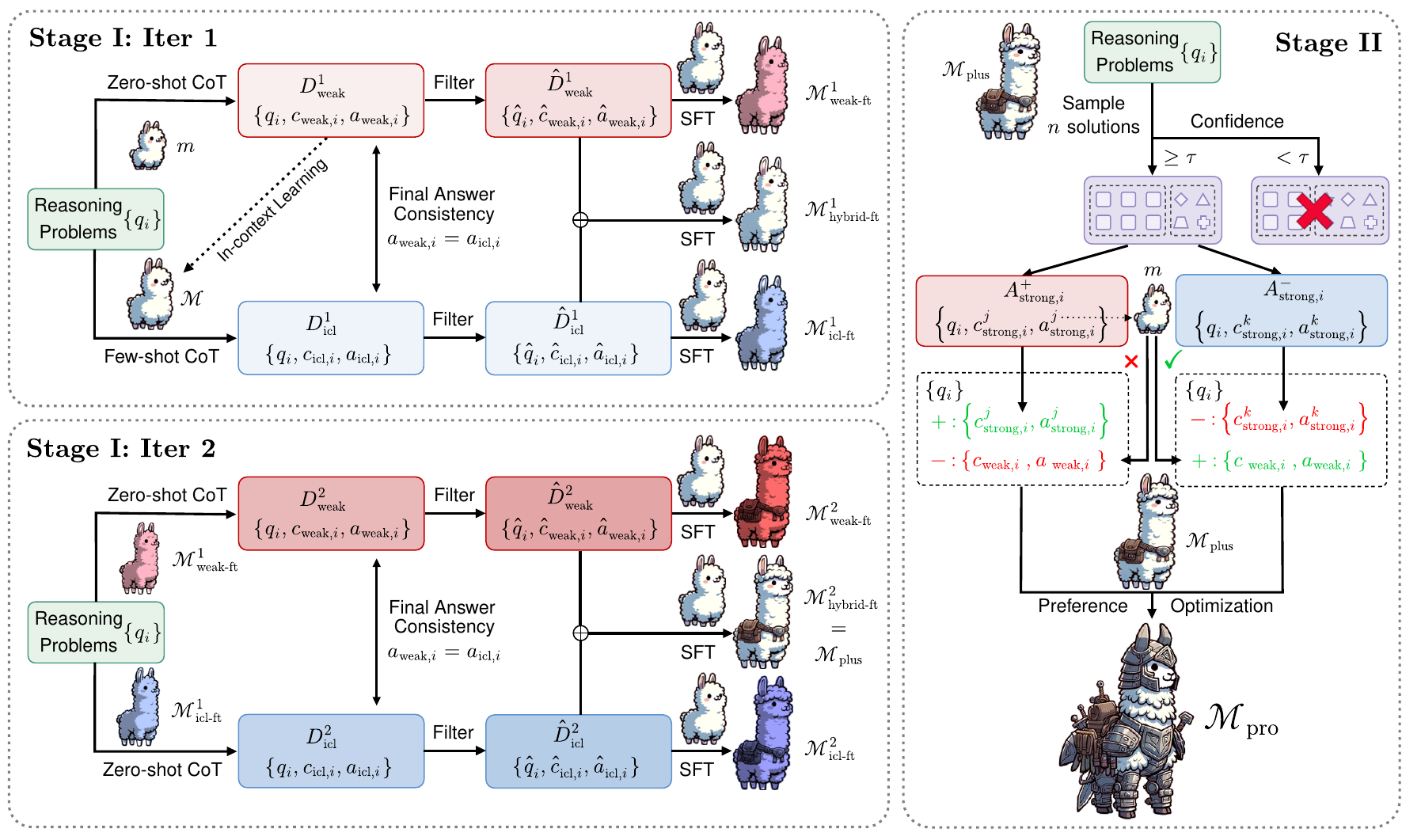}
    \vspace{-6mm}
    \caption{
    \small \textbf{Overview of our method evolving from $\M$ \raisebox{-.2ex}{\includegraphics[height=2ex]{figures/llama_bigm.png}} $\to$ $\M_\text{plus}$ \raisebox{-.2ex}{\includegraphics[height=2.5ex]{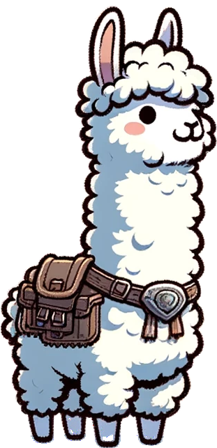}} $\to$ $\M_\text{pro}$ \raisebox{-.2ex}{\includegraphics[height=3ex]{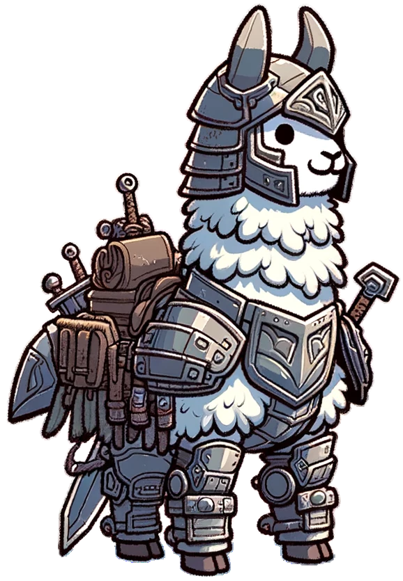}}.}
    \textbf{Left:} we utilize \textit{final answer consistency} to selectively filter weak and icl data from diverse sources, which is used to fine-tune the strong model $\M$ and obtain $\M_\text{plus}$ with enhanced mathematical reasoning capabilities. \textbf{Right:} we leverage the confidence of $\M_\text{plus}$ to identify contrastive samples for performance optimization, resulting in a more robust strong model $\M_\text{pro}$.
    }
    \label{fig:fig2}
    \vspace{-3mm}
\end{figure*}

In this section, we propose a weak-to-strong training method designed to maximize the use of weak data and to elicit the strong model's innate talent. First, we identify potentially positive samples in the absence of ground truth and external signals. During Stage I, we exclusively utilize this subset of data for supervised fine-tuning. Then once the strong model has achieved a certain level of reasoning proficiency, we employ the full weak data, particularly the potentially negative samples in Stage II via preference learning-based approaches like DPO~\cite{DBLP:conf/nips/RafailovSMMEF23}, encouraging the strong model to learn from mistakes made by the weaker model. The whole framework is depicted in Fig.~\ref{fig:fig2}.

\subsection{Stage I: Learn from ``Positive'' Samples}
\label{sec:sft}
Given a weak model $m$ and a series of math problems $\mathcal{Q}$ \textbf{without} ground truth, $m$ generates weak data $\D_\text{weak} = \{q_i, c_{\text{weak}, i}, a_{\text{weak}, i}\}$, where $q_i \in \mathcal{Q}$, $c_{\text{weak}, i}$ represents a reasoning chain, and $a_{\text{weak}, i}$ represents the final answer. The correctness of $a_{\text{weak}, i}$ is unknown. 
The central challenge is: \textbf{how can we maximize the use of $m$ and $\D_\text{weak}$ to fully enhance and recover the mathematical reasoning capabilities of a stronger model $\M$?}

\subsubsection{Full Weak Fine-Tuning}
Our initial strategy is to fine-tune the stronger model $\M$ across the entirety of the weak dataset $\D_\text{weak}$. While prior research \citep{DBLP:journals/corr/abs-2312-09390} has validated the effectiveness of this approach in text classification tasks, its efficacy in reasoning tasks remains unexplored. We have therefore embarked on an investigation to determine whether the phenomenon of \textit{weak-to-strong generalization} can also enhance the reasoning capabilities of $\M$ in this less examined domain.

\subsubsection{Weak In-Context Learning}
Another straightforward approach is in-context learning (ICL, \citet{DBLP:journals/corr/abs-2301-00234}), which requires only several training samples as demonstrations in the prompt. Specifically, we randomly select four samples from $\D_\text{weak}$ as demonstrations. Since we do not have access to the ground truth, these demonstrations cannot be provably correct.

\subsubsection{Weak-ICL Fine-Tuning}
Given that models can mimic weak errors through supervised fine-tuning \citep{charikar2024quantifying,lang2024theoretical}, we propose refining $\D_\text{weak}$ before use, instead of using all data blindly. Additionally, we seek to harness the innate abilities of the strong model activated via in-context learning. Building on these two ideas, we introduce \textit{weak-icl fine-tuning}, employing both weak data $\D_\text{weak}$ and ``icl data'' $\D_\text{icl} = \{q_i, c_{\text{icl}, i}, a_{\text{icl}, i}\}$,  where $q_i \in \mathcal{Q}$, $c_{\text{icl}, i}$ and $a_{\text{icl}, i}$ are generated by $\M$ with few-shot demonstrations,\footnote{Experiments in \S\ref{sec:icl_performance} show that despite ICL being affected by demonstration selection, our method can achieve further improvements accordingly beyond ICL.} as higher-quality supervision signals.

Note that, for both $\D_\text{weak}$ and $\D_\text{icl}$, we cannot determine whether a certain answer is correct or not. Nonetheless, when two models, employing distinct data representations, converge on the same answer in an open-ended task, it is indicative of a higher likelihood of accuracy. This phenomenon supports the reliability of the results when consistency is observed across different methodologies. We thus compare $\D_\text{weak}$ and $\D_\text{icl}$ generated by the weak model and strong model, respectively, and select $\hat{\D}_\text{weak}$ and $\hat{\D}_\text{icl}$ if $a_{\text{weak}, i} = a_{\text{icl}, i}$, for subsequent supervised fine-tuning. We call this approach \textit{final answer consistency}. Considering the combination of the two sets of data, we can obtain three versions of enhanced fine-tuned strong models:
\begin{itemize*}
    \item $\M_\text{weak-ft}$: $\M$ fine-tuned on $\hat{\D}_\text{weak}$.
    \item $\M_\text{icl-ft}$: $\M$ fine-tuned on $\hat{\D}_\text{icl}$.
    \item $\M_\text{hybrid-ft}$: $\M$ fine-tuned on the union of $\hat{\D}_\text{weak}$ and $\hat{\D}_\text{icl}$.
\end{itemize*}

\paragraph{Iterative Training}
Upon closed examination of $\M_\text{weak-ft}$ and $\M_\text{icl-ft}$, we see that they still satisfy the condition of having different data representations, as they are trained on data from different sources—$\hat{\D}_\text{weak}$ is generated by the weak model, whereas $\hat{\D}_\text{icl}$ primarily originates from the strong model itself. Hence, we can perform iterative training to bootstrap performance. We denote the initial round of supervised fine-tuning data as $\hat{\D}_\text{weak}^1$ and $\hat{\D}_{\text{icl}}^1$, resulting in models $\M_\text{weak-ft}^1$, $\M_\text{icl-ft}^1$, and $\M_\text{hybrid-ft}^1$. In the second iteration, we obtain zero-shot solutions from $\M_\text{weak-ft}^1$ applied to $\Q$ to construct $\D_{\text{weak}}^2$, and those from $\M_\text{icl-ft}^1$ to construct $\D_{\text{icl}}^2$. Here, the subscripts ``weak'' and ``icl'' indicate the initial data source. Then we apply final answer consistency to obtain $\hat{\D}_{\text{weak}}^2$ and $\hat{\D}_{\text{icl}}^2$. Following another round of supervised fine-tuning, we have:
\begin{itemize*}
    \item $\M_\text{weak-ft}^2$: $\M$ fine-tuned on $\hat{\D}_\text{weak}^2$.
    \item $\M_\text{icl-ft}^2$: $\M$ fine-tuned on $\hat{\D}_\text{icl}^2$.
    \item $\M_\text{hybrid-ft}^2$: $\M$ fine-tuned on the union of $\hat{\D}_\text{weak}^2$ and $\hat{\D}_\text{icl}^2$.
\end{itemize*}

Note that the iterative training step is optional; it may lead to performance degradation when data quality is too low or the model overfits.

\subsection{Stage II: Learn from ``Negative'' Samples}
\label{sec:dpo}

We denote the final iteration of $\M_{\text{hybrid-ft}}$ from Stage I as $\M_{\text{plus}}$, which has learned dual mathematical solutions and holds potential for further enhancement. Next, we apply preference optimization techniques to strategically utilize the potentially erroneous subset of the original weak dataset $\D_\text{weak} = \{q_i, c_{\text{weak}, i}, a_{\text{weak}, i}\}$ generated by $m$, which allows the strong model to identify and avoid similar errors in future reasoning processes. The key factor lies in how to construct contrastive samples for learning.

\begin{table}[t]
    \centering
    \small
    \begin{tabular}{p{0.95\columnwidth}}
        \toprule
        \vspace{-2mm}
        \textbf{\textsc{Question}} ($q_i$): John has five more roommates than \textbf{twice} as many as Bob. If Bob has 10 roommates, how many roommates does John have?\\
        \midrule
        \textcolor{red}{\textbf{\textsc{Weak Response}} ($\{c_{\text{weak}, i}, a_{\text{weak}, i}\}$):} John has 10+5=\textcolor{red}{15} roommates. The answer is \textcolor{red}{15}.\\
        \midrule
        \textcolor{GreenCheck}{\textbf{\textsc{Self Response 1}} ($\{c_{\text{strong}, i}^1, a_{\text{strong}, i}^1\} \in A_{\text{strong}, i}^+$):} Bob has 10 roommates. Twice as many as Bob is 2*10 = 20 roommates. John has 5 more roommates than twice as many as Bob, so John has 20+5 = 25 roommates. The answer is \textcolor{GreenCheck}{25}.\\
        \vspace{1mm}
        \textcolor{GreenCheck}{\textbf{\textsc{Self Response 2}} ($\{c_{\text{strong}, i}^2, a_{\text{strong}, i}^2\} \in A_{\text{strong}, i}^+$):} Let x be the number of roommates Bob has. John has 5 more roommates than twice as many as Bob, so John has 2x+5 roommates. Bob has 10 roommates, so x=10. John has 2*10+5 = 25 roommates. The answer is \textcolor{GreenCheck}{25}.\\
        \bottomrule
    \end{tabular}
    \caption{\small A real case example. Given a math question, the \textcolor{red}{incorrect} ``weak response'' is generated by $m$, while the two \textcolor{GreenCheck}{correct} ``self responses'' are sampled from $A_{\text{strong}, i}^+$ self-generated by $\M_{\text{plus}}$. Benefiting from dual solutions in the training data during Stage I, $\M_{\text{plus}}$ is able to generate different reasoning paths that converge to the same final answer. Through Stage II, $\M_{\text{plus}}$ learns to avoid $m$'s error of overlooking the key word ``\textbf{twice}'' in calculations.}
    \label{tab:example}
    \vspace{-3mm}
\end{table}

Without access to ground truth, the current strong model with enhanced reasoning capabilities identifies the most likely correct answers based on its confidence. Specifically, for each question $q_i \in \Q$, we sample $n$ responses from $\M_\text{plus}$, and define the probability of the answer that appears most frequently among these responses as \textit{confidence}. When the confidence falls below a specified threshold $\tau$, we consider the model's judgment on this question unreliable and therefore discard it. Conversely, if the confidence is no less than $\tau$, we regard the model as capable of solving the question and proceed to construct contrastive samples as follows.
\begin{itemize*}
    \item For a question $q_i$ where $\M_\text{plus}$ is confident, we denote the most confident answer as $a_{\text{strong}, i}^+$ and $P(a_{\text{strong}, i}^+) \ge \tau$. It can be considered as the ``correct'' answer according to $\M_\text{plus}$. For instance, if we set $\tau=0.6$ and 8 out of 10 sampled responses have the same final answer ``4.2'', we say that $\M_\text{plus}$ considers ``4.2'' to be the correct answer to this question, i.e. $a_{\text{strong}, i}^+ = 4.2$.
    \item Then we divide the sampled $n$ responses of $\M_\text{plus}$ to $q_i$ into two sets: $A_{\text{strong}, i}^+ = \{c_{\text{strong}, i}^j, a_{\text{strong}, i}^j\}$ where $a_{\text{strong}, i}^j = a_{\text{strong}, i}^+$; $A_{\text{strong}, i}^- = \{c_{\text{strong}, i}^k, a_{\text{strong}, i}^k\}$ where $a_{\text{strong}, i}^k \ne a_{\text{strong}, i}^+$. In the above example, $|A_{\text{strong}, i}^+| = 8$ and $|A_{\text{strong}, i}^-| = 2$.
    \item If the weak model holds an answer that the enhanced model considers ``correct'', that is, $a_{\text{weak}, i} = a_{\text{strong}, i}^+$, we treat the weak model's response $\{c_{\text{weak}, i}, a_{\text{weak}, i}\}$ as chosen response and randomly select a rejected response from $A_{\text{strong}, i}^-$. Otherwise, if $a_{\text{weak}, i} \ne a_{\text{strong}, i}^+$, we treat $\{c_{\text{weak}, i}, a_{\text{weak}, i}\}$ as rejected response and randomly select a chosen response from $A_{\text{strong}, i}^+$. Examples are shown in Tab.~\ref{tab:example}.
\end{itemize*}

Further training $\M_\text{plus}$ on these samples enables it to distinguish between correct and incorrect solutions, leading to a stronger model $\M_\text{pro}$.

\begin{figure*}[t]
    \centering
    \includegraphics[width=0.95\linewidth]{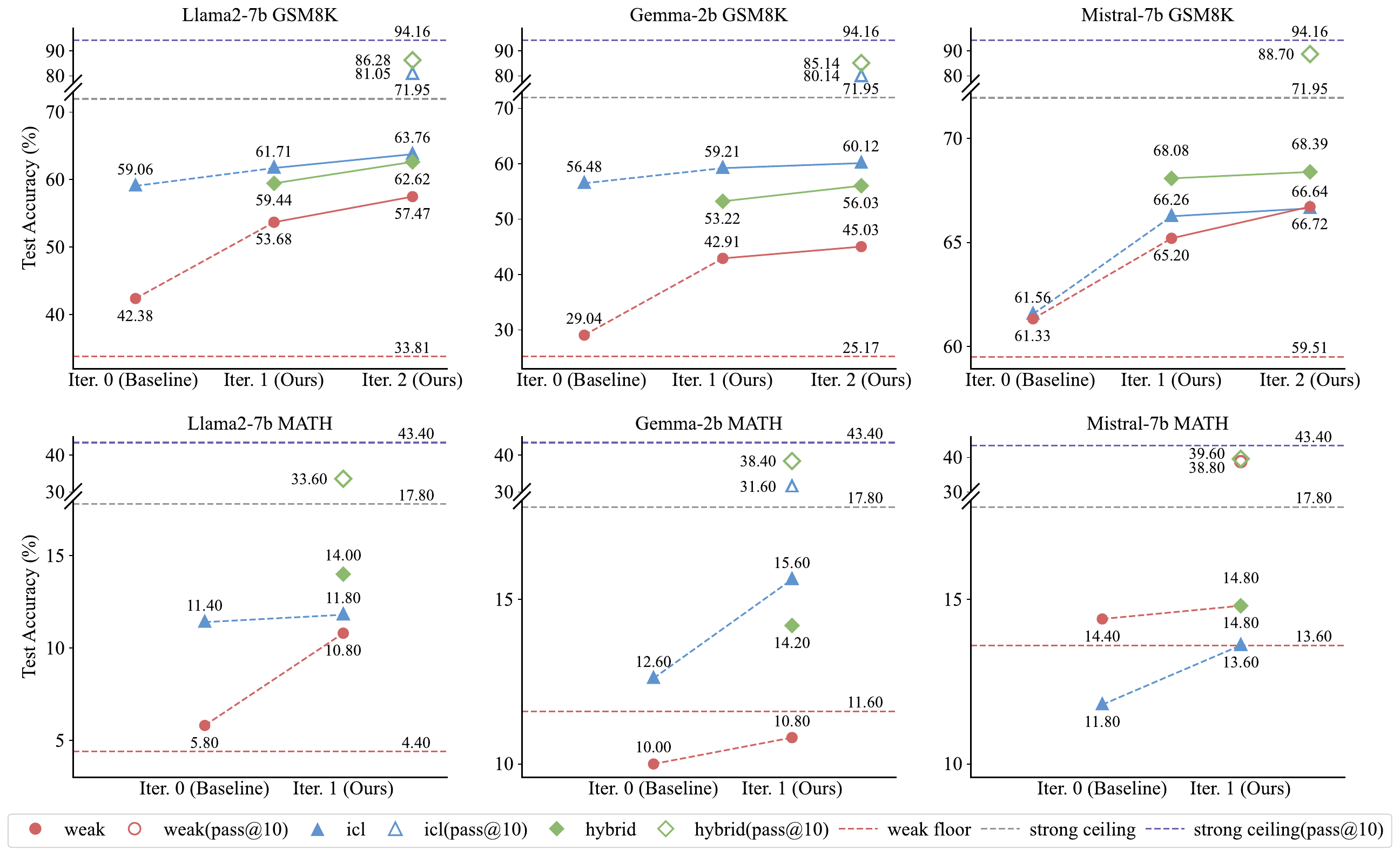}
    \caption{\small Main results of Stage I. ``Iter. 0'' presents the performance of two baselines, where ``\textcolor{weak}{weak}'' indicates full weak fine-tuning, i.e., naively fine-tuning on the entire weak data, and ``\textcolor{icl}{icl}'' refers to weak ICL without fine-tuning. Models connected by a line mean that they share the same training data sources. Results below ``\textcolor{strongceiling}{strong ceiling}'' present test accuracy via greedy decoding, while those above show pass@k scores ($k=10$ and $\text{temperature}=1.0$). For simplicity, we only present the pass@k scores of $\M_\text{hybrid-ft}$ and checkpoints that surpass it using greedy decoding, and full results are provided in \S\ref{sec:passk}.}
    \label{fig:main_results}
    \vspace{-2mm}
\end{figure*}

\section{Experiments}

\subsection{Datasets}

\begin{wraptable}{r}{0.50\textwidth}
    \vspace{-3mm}
    \centering
    \small
    \begin{tabular}{l@{\hspace{25pt}}c@{\hspace{25pt}}c@{\hspace{25pt}}c}
    \toprule
     & \textbf{\# $\D_{\text{gold}, 1}$} & \textbf{\# $\D_{\text{gold}, 2}$} & \textbf{\# Test} \\
    \midrule
    GSM8K & 7,000 & 7,000 & 1,319 \\
    MATH & 6,000 & 6,000 & 500 \\
    \bottomrule
    \end{tabular}
    \caption{\small Data Statistics. $\D_{\text{gold}, 1}$ and $\D_{\text{gold}, 2}$ are subsets of the training set. The weak model uses $\D_{\text{gold}, 1}$ to cultivate initial mathematical skills, while the strong model can only access questions from $\D_{\text{gold}, 2}$ without ground truths.}
    \label{tab:data_stat}
    \vspace{-3mm}
\end{wraptable}

GSM8K \citep{DBLP:journals/corr/abs-2110-14168} and MATH \citep{DBLP:conf/nips/HendrycksBKABTS21} are two widely used datasets for mathematical reasoning, and MATH comprises more challenging competition problems. The data statistics we use are presented in Tab.~\ref{tab:data_stat}. Particularly, to ensure a sufficient amount of training data for developing preliminary mathematical skills in the weak model, we augment the GSM8K training set with the data constructed by \citet{abel}. Further details are available in \S\ref{sec:dataset_details}.

\subsection{Experiment Settings}
We use Llama2-70b as the strong model and employ three weak models from different families: Llama2-7b, Gemma-2b, and Mistral-7b. We apply full parameter fine-tuning to the weak models on $\D_{\text{gold}, 1}$, and consistently adopt LoRA \citep{DBLP:conf/iclr/HuSWALWWC22} to fine-tune the strong model. In Stage I, we perform two rounds of iterations on GSM8K and one round on MATH according to the principles of iteration outlined in \S\ref{sec:sft}. In Stage II, we adopt two preference learning-based approaches, DPO \citep{DBLP:conf/nips/RafailovSMMEF23} and its variant ORPO \citep{DBLP:journals/corr/abs-2403-07691}. Details are provided in \S\ref{sec:training_details}.

We evaluate the accuracy on the test set. The performance of the weak model $m$ is defined as the ``\textbf{weak floor}''. The performance of the strong model $\M$, fine-tuned with data containing gold solutions from $\D_{\text{gold}, 2}$, is termed the ``\textbf{strong ceiling}''. It represents the upper limit of the capabilities that the strong model can achieve with $\D_{\text{gold}, 2}$.

\subsection{Results of Stage I}
\label{sec:stagei_results}
The main results of Stage I on both GSM8K and MATH datasets are depicted in Fig.~\ref{fig:main_results}.\footnote{We do not incorporate the zero-shot performance of the strong model in the main results as it is significantly lower than that of weak ICL. See \S\ref{sec:zero-shot} for further details.} Notably, in the MATH experiments, we randomly sample additional data that is not chosen based on the final answer consistency, due to the small amount available. Please refer to \S\ref{sec:stagei_math} for details. According to Fig.~\ref{fig:main_results}, we have the following observations.

\textbf{Weak-ICL fine-tuning demonstrates a notable enhancement.}
Using our proposed method, the performance of the strong model, supervised \textit{only} by the weak Gemma-2b with 25.17 accuracy on GSM8K (without any gold answers), can be improved up to 60.12, surpassing naive full weak fine-tuning by 31.08, and $\M_\text{plus}$ (i.e., $\M_{\text{hybrid-ft}}^2$) outperforms it by 26.99. This verifies the effectiveness of data refining before supervised fine-tuning. Also, experimental results show that the mathematical reasoning capabilities of the strong model are increasingly recovered as the weak model improves, a conclusion verified by \citet{DBLP:journals/corr/abs-2402-15505} on classification tasks. In detail, the performance on GSM8K gradually improves for Gemma-2b, Llama-7b, and Mistral-7b ($25.17 \to 33.81 \to 59.51$). Hence, the maximum performance of the strong model, trained with data generated by these models, also progressively enhances ($60.12 \to 63.76 \to 68.39$).

\textbf{$\M_{\text{hybrid-ft}}$ achieves the highest pass@k scores.} As expected, $\M_{\text{hybrid-ft}}$ achieves the highest pass@k scores in the weak-to-strong setting, benefiting from its training data that incorporates two types of solutions—one from the weak model, and another from the strong model. This diversity enhances the robustness of the model by reducing the likelihood of overfitting. Additionally, the performance of $\M_\text{icl-ft}$ generally surpasses that of $\M_\text{weak-ft}$, which can be attributed to variations in process-level accuracy and possibly the solution format. Detailed analyses are conducted in \S\ref{sec:appendix_analysis}. 

\textbf{Naive fine-tuning is inadequate for weak-to-strong reasoning.} When using Gemma-2b as the weak model on the MATH dataset, full weak fine-tuning underperforms compared to the weak floor (10.0 v.s. 11.6). This indicates that naive fine-tuning, though successfully applied to classification, chess, and reward modeling tasks \citep{DBLP:journals/corr/abs-2312-09390},  falls short for intricate reasoning tasks, particularly those of substantial difficulty like questions in MATH. In contrast, our weak-icl fine-tuning method effectively bridges the gap, offering an effective and scalable solution for the weak-to-strong reasoning challenge.

\paragraph{Effect of ICL Performance}
\label{sec:icl_performance}

\begin{wrapfigure}{r}{0.5\textwidth}
    \centering
    \begin{minipage}{0.5\textwidth}
    \vspace{-3mm}
    \centering
    \includegraphics[width=0.75\linewidth]{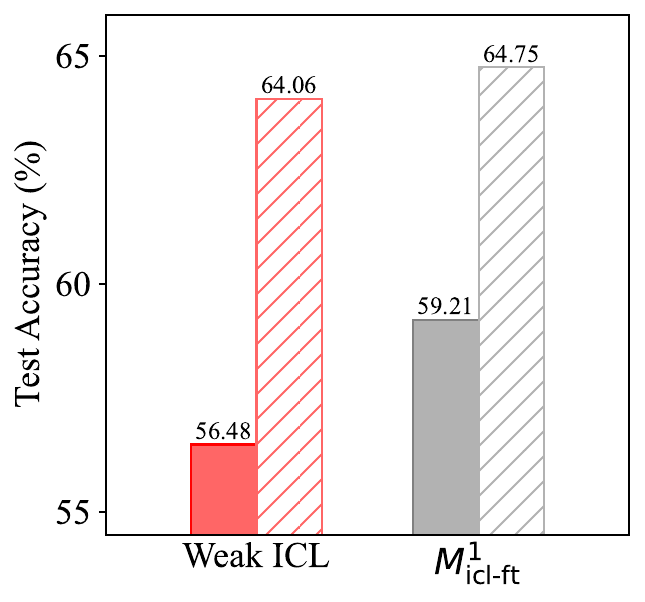}
    \vspace{-1mm}
    \caption{\small Results on GSM8K supervised by Gemma-2b. \raisebox{-.2ex}{\includegraphics[height=1.5ex, width=2.5ex]{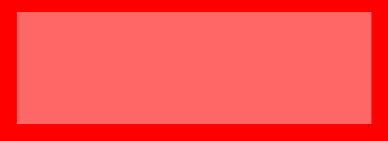}} and \raisebox{-.2ex}{\includegraphics[height=1.5ex, width=2.5ex]{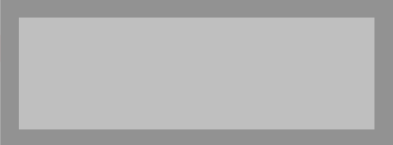}} are under original demonstrations, and \raisebox{-.2ex}{\includegraphics[height=1.5ex, width=2.5ex]{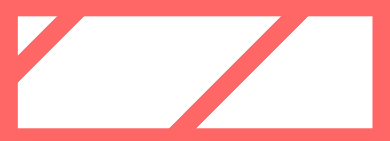}} and \raisebox{-.2ex}{\includegraphics[height=1.5ex, width=2.5ex]{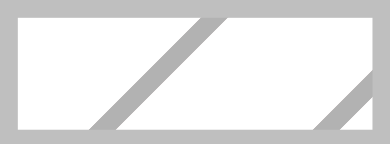}} are under carefully selected demonstrations.}
    \label{fig:icl_performance}
    \end{minipage}
\end{wrapfigure}

Given that the efficacy of weak-icl fine-tuning partially depends on the effectiveness of weak ICL, we further explore how enhancing ICL performance through careful selection of demonstrations affects the performance of weak-icl fine-tuning. Fig.~\ref{fig:icl_performance} shows the test accuracy on GSM8K using Gemma-2b as the weak model under a different set of demonstrations.

The results indicate that the performance of weak ICL with this particular group of demonstrations increases from the original 56.48 to 64.06. We then regenerate $\D_\text{icl}$ with these demonstrations in the prompt and fine-tune the strong model on $\hat{\D}_{\text{icl}}$, which is selectively curated through final answer consistency. This further improves performance from 64.06 to 64.75, confirming the utility of self-directed data curation. It is worth noting that although weak ICL holds the potential for high performance, the selection of effective demonstrations in a weak-to-strong framework is a non-trivial thing, and is beyond the scope of this paper.

\subsection{Results of Stage II}
\label{sec:stageii_results}

\begin{wraptable}{r}{0.5\textwidth}
    \vspace{-3mm}
    \centering
    \small
    \begin{tabular}{lccc}
        \toprule
        \multirow{2}{*}{\textbf{Weak Model}} & \multicolumn{3}{c}{\textbf{Test Accuracy}} \\
        \cmidrule(lr){2-4}
         & I & II. DPO & II. ORPO \\
        \midrule
        \multicolumn{3}{l}{\textbf{GSM8K}} \\
        Llama2-7b & 62.62 & 66.19 (\textcolor{GreenCheck}{+3.57}) &  68.16 (\textcolor{GreenCheck}{+5.54}) \\
        Gemma-2b & 56.03 & 64.52 (\textcolor{GreenCheck}{+8.49}) & 63.91 (\textcolor{GreenCheck}{+7.88}) \\
        Mistral-7b & 68.39 & 70.96 (\textcolor{GreenCheck}{+2.57}) & 72.18 (\textcolor{GreenCheck}{+3.79}) \\
        \midrule
        \multicolumn{3}{l}{\textbf{MATH}} \\
        Llama2-7b & 14.00 & 12.00 (\textcolor{red}{-2.00}) & 15.00 (\textcolor{GreenCheck}{+1.00}) \\
        Gemma-2b & 14.20 & 11.60 (\textcolor{red}{-2.60}) & 16.00 (\textcolor{GreenCheck}{+1.80}) \\
        Mistral-7b & 14.80 & 13.40 (\textcolor{red}{-1.40}) & 17.00 (\textcolor{GreenCheck}{+2.20}) \\
        \bottomrule
    \end{tabular}
    \caption{\small Main results of Stage II.}
    \label{tab:dpo_results}
    \vspace{-3mm}
\end{wraptable}

As discussed in \S\ref{sec:dpo}, we employ the final iteration of $\M_{\text{hybrid-ft}}$ as $\M_{\text{plus}}$ for subsequent preference learning. The experimental results in \S\ref{sec:stagei_results} validate this checkpoint achieves higher pass@k and possesses significant potential for further refinement.

As shown in Tab.~\ref{tab:dpo_results}, our method for constructing positive and negative samples effectively enhances the strong model's math reasoning capabilities. On GSM8K, both DPO and ORPO consistently achieve significant improvements using our constructed datasets, notably resulting in an increase of 8.49 points when supervised by Gemma-2b. Despite the inherently challenging nature of MATH problem, which compromises the strong model's judgment and introduces inaccuracies in the training data, our method still achieves improvements on MATH through ORPO by at least 1 point.\footnote{\citet{DBLP:journals/corr/abs-2404-19733,DBLP:journals/corr/abs-2404-02893,DBLP:journals/corr/abs-2404-02078} demonstrate that DPO can cause performance degradation on MATH due to the lack of regularization in its loss.}

\paragraph{Data Construction Recipe}

When constructing preference data, we always use weak responses generated by the weak model as one of the chosen/rejected responses, instead of relying exclusively on self-generated data. We also test the self-generated setting on GSM8K using Llama2-7b as the weak model, where both chosen and rejected responses are generated by the strong model itself. The DPO test accuracy in this setting is 62.40 (\textcolor{red}{-0.22}), indicating a slight performance degradation. Without ground truth, the constructed positive and negative samples actually correspond to the more frequently and less frequently occurring answers, respectively, and are related to the answers the model tends to choose. Since preference optimization essentially performs ranking, the potential benefit of this self-generated setting is minimal. Therefore, incorporating weak data signals in the preference data construction process proves to be a better approach.

\subsection{Analysis}

\begin{figure*}[t]
    \centering
    \includegraphics[width=0.95\linewidth]{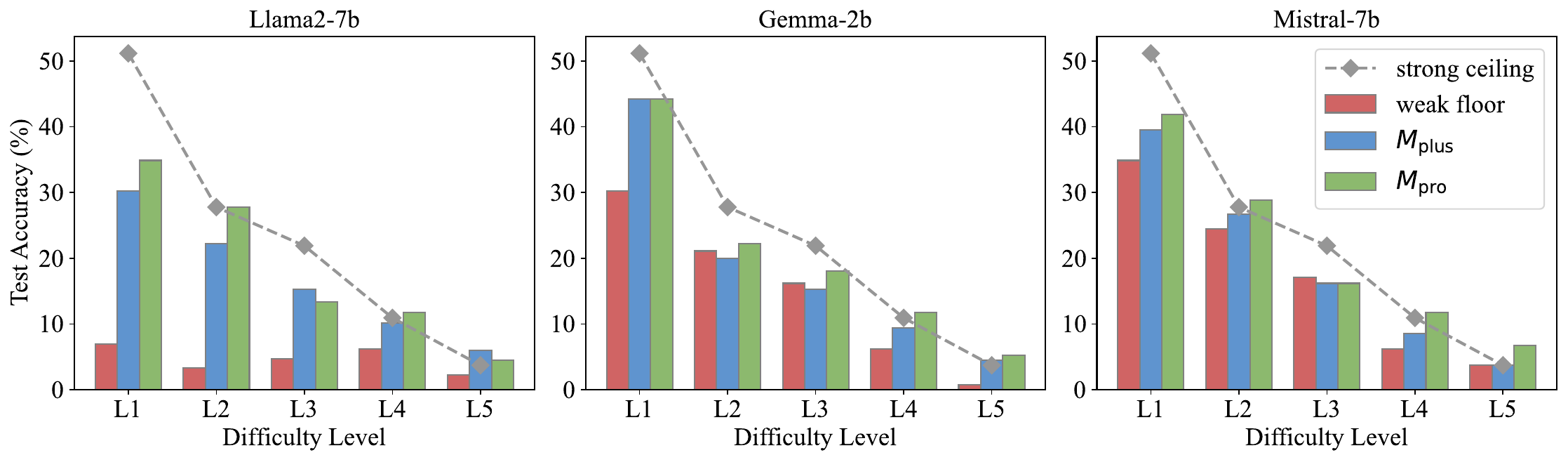}
    \caption{\small Test accuracy across varying difficulty levels on the MATH test set. We use ORPO to obtain $\M_{\text{pro}}$.}
    \label{fig:analysis}
    \vspace{-2mm}
\end{figure*}

For further analysis, we examine the accuracy across different difficulty levels in the MATH test set (See \S\ref{sec:math_data} for data statistics).

As shown in Fig.~\ref{fig:analysis}, the strong model exhibits better generalization on easier problems. Specifically, even though Llama2-7b achieves only 6.98 points accuracy on level 1 problems, Llama2-70b can achieve an accuracy exceeding 30 points after training using this weak supervision. For more challenging problems (levels 4-5), $\M_\text{pro}$, enhanced with ORPO, even \textbf{surpasses the strong ceiling obtained by supervised fine-tuning solely on gold solutions}. This phenomenon serves to validate the effectiveness of learning from incorrect data.

\subsection{Experiments Closer to Future Scenarios}

\begin{wraptable}{r}{0.4\textwidth}
    \vspace{-3mm}
    \centering
    \small
    \begin{tabular}{lcc}
        \toprule
         & \textbf{Test Accuracy} \\
        \midrule
        Weak Floor & 11.82 \\
        Full Weak FT & 12.46 \\
        Weak ICL & 8.63 \\
        \midrule
        $\M_{\text{weak-ft}}^1$ & 12.78 \\
        $\M_{\text{icl-ft}}^1$ & 9.58 \\
        $\M_{\text{hybrid-ft}}^1$ & 11.18 \\
        $\M_{\text{weak-ft}}^2$ & \underline{13.10} \\
        $\M_{\text{icl-ft}}^2$ & 11.50 \\
        $\M_{\text{hybrid-ft}}^2$ ($\M_{\text{plus}}$) & 11.82 \\
        \midrule
        $\M_{\text{pro}}$ & \textbf{15.65} \\
        \bottomrule
    \end{tabular}
    \caption{\small Results on \emph{OlympicArena} using \emph{Llama3 family}. The best result is in \textbf{bold}, and the best result of supervised fine-tuning is \underline{underlined}.}
    \label{tab:olympic_results}
    % \vspace{-3mm}
\end{wraptable}

In preliminary tests with Llama3-70b \citep{llama3modelcard}, we observe that on GSM8K and MATH, Llama3-70b can largely unlock its potential through in-context learning, with marginal or even adverse impacts from parameter updates due to training instabilities. Consequently, we focus on a more challenging dataset developed after the release of Llama3-70b, OlympicArena \citep{huang2024olympicarenabenchmarkingmultidisciplinecognitive}, to simulate a more realistic future scenario.

We only consider English questions in OlympicArena, excluding the CODE (Code Generation) and OT (Others) problem types that require case-based or expert evaluation. This results in 6,020 training data without solutions and final answers, and 313 test data with final answers to assess the performance of different methods. We use Llama3-8b-instruct (without initial fine-tuning on a subset of training data) as the weak model and Llama3-70b as the strong model to be improved. This configuration more closely resembles future real-world weak-to-strong scenarios.

Experimental results are displayed in Tab.~\ref{tab:olympic_results}. ``Weak Floor'' represents the zero-shot performance of Llama3-8b-instruct, ``Full Weak FT'' denotes the performance of Llama3-70b after supervised fine-tuning on the full set (i.e, 6,020) of weak solutions generated by Llama3-8b-instruct on the training set, and ``Weak ICL'' indicates the performance of Llama3-70b under 4-shot weak demonstrations generated by Llama3-8b-instruct. Despite having more parameters, Llama3-70b under in-context learning still performs lower than the zero-shot performance of Llama3-8b-instruct due to insufficient mining capabilities.

$\M_{\text{weak-ft}}^1$, obtained by our proposed weak-icl fine-tuning method, achieves higher performance than Full Weak FT with fewer training data (i.e., 746), outperforming it by 0.32 points. After the second stage of preference optimization, which further exploits the weak model and training questions without answers, the strong model's performance is improved by an additional 3.19 points over Full Weak FT. This demonstrates the robustness and generalizability of our method in scenarios closer to future conditions.

\section{Related Work}
\subsection{LLM Training}
LLMs can enhance their ability to solve tasks and better align with human instructions through a supervised fine-tuning (SFT) phase \citep{DBLP:journals/corr/abs-2308-10792,DBLP:journals/corr/abs-2310-05492,DBLP:journals/corr/abs-2306-09782,DBLP:journals/corr/abs-2310-10195}. This phase heavily relies on the quality of training data, as previous studies \citep{DBLP:conf/nips/ZhouLX0SMMEYYZG23,DBLP:conf/acl/WangKMLSKH23} demonstrate that higher data quality translates to substantial gains in model performance. In this paper, we investigate the potential of learning from weak supervision.

To further align LLMs with human values and enable learning from both positive and negative feedback, additional training is required, such as reinforcement learning from human feedback (RLHF, \citet{DBLP:conf/nips/Ouyang0JAWMZASR22,DBLP:journals/corr/abs-2212-08073}) and direct preference optimization (DPO, \citet{DBLP:conf/nips/RafailovSMMEF23}). In particular, DPO reparameterizes reward functions in RLHF and has been widely used due to its simplicity. Several variants of DPO have then emerged to further enhance its stability and performance, such as ORPO \citep{DBLP:journals/corr/abs-2403-07691} and SimPO \citep{meng2024simpo}, etc. This paper explores the capabilities of DPO and ORPO using our constructed contrastive samples in a weak-to-strong setting.

\subsection{Mathematical Reasoning}
The exploration of mathematical reasoning within LLMs has been a focal point for evaluating their cognitive capabilities akin to human reasoning \citep{DBLP:conf/acl/QiaoO0CYDTHC23,DBLP:conf/acl/Lu00WC23}. Researchers have developed various methods to enhance the mathematical reasoning capabilities of LLMs after pre-training, which can be broadly classified into two categories: (1) Prompting: Some works \citep{DBLP:conf/nips/KojimaGRMI22,DBLP:conf/nips/Wei0SBIXCLZ22,DBLP:conf/iclr/ZhouSHWS0SCBLC23,DBLP:conf/emnlp/LiuG0HZQZ23} aim to elicit the intrinsic reasoning abilities of LLMs by specific prompting engineering, without updating the model parameters; (2) Fine-tuning: Another line of studies focuses on generating a more extensive and higher-quality collection of question-answer pairs \citep{DBLP:journals/corr/abs-2309-12284,DBLP:journals/corr/abs-2312-17120,DBLP:journals/corr/abs-2312-08935}. Through supervised fine-tuning and preference optimization \citep{DBLP:journals/corr/abs-2308-09583,DBLP:journals/corr/abs-2310-10631,DBLP:journals/corr/abs-2402-14830,DBLP:journals/corr/abs-2404-02893}, the models can achieve significant improvements in their mathematical problem-solving capabilities.

\section{Conclusion}
In this paper, we explore the efficacy of weak-to-strong framework in complex reasoning tasks. We introduce a new method that elicits strong capabilities using weak supervision, without relying on annotations from humans or more advanced models. This method focuses on the strong model's ability to autonomously refine its training data, even if it has not learned the task before. By iteratively expanding its learning scope, the strong model continuously broadens its reasoning skills. This self-directed data curation is crucial for scaling up the enhancement of AI reasoning capabilities, making the model more independent and effective in its developmental trajectory. Through this work, we seek to illuminate new pathways for AI development, emphasizing the critical role of innovative model supervision in advancing AGI and beyond.

\section*{Limitations}
In our experiments, we use Llama2-70b and Llama3-70b as a proxy for hypothetical superintelligent models of the future. We acknowledge that there might be performance discrepancies compared to a genuine future advanced model. Nonetheless, our efforts lay the groundwork for investigating methodologies in weak-to-strong reasoning. Additionally, this paper does not explore supervision at the process level, such as the model's ability to learn from partially correct data \citep{DBLP:conf/iclr/NiIWPMRG23,DBLP:journals/corr/abs-2305-20050}. In the weak-to-strong scenario, the presence of non-negligible errors and noise at the process level yields only limited performance improvements in our early experiments, thereby posing challenges for future research.

\section*{Acknowledgements}
We sincerely thank Xuefeng Li, Haoyang Zou, and Ting Wu for their valuable insights during discussions, which greatly enhanced the quality of this work. This work was supported by Shanghai Artificial Intelligence Laboratory, SJTU SEIEE - ByteDance Large Language Model Joint Laboratory.

\bibliography{main}
\bibliographystyle{acl_natbib}

\clearpage
\appendix

\section{Appendix}
\label{sec:appendix}

\subsection{Dataset Details}
\label{sec:dataset_details}

\subsubsection{Dataset Construction}
For GSM8K, we evenly divide the original training dataset of 7,473 samples into two subsets, $\D_{\text{gold}, 1}$ and $\D_{\text{gold}, 2}$. Additionally, we supplement both $\D_{\text{gold}, 1}$ and $\D_{\text{gold}, 2}$ with the data of the same distribution developed by \citep{abel}, until each contains 7,000 samples. Thus, the weak model uses $\D_{\text{gold}, 1}$, which includes both questions and gold solutions, to obtain basic problem-solving capabilities. Meanwhile, the strong model can only access a training dataset $\Q = \{q_i\}$, where $q_i \in \D_{\text{gold}, 2}$, consisting of 7,000 mathematical problems without ground truth answers. GSM8K also includes 1,319 test samples.

For MATH, we employ the same subset of 500 representative problems as the test set, identical to that used in \citet{DBLP:journals/corr/abs-2305-20050}. We then split the remaining 12,000 samples evenly between $\D_{\text{gold}, 1}$ and $\D_{\text{gold}, 2}$, each containing 6,000 samples.

\subsubsection{Statistics of MATH test set}
\label{sec:math_data}
\begin{wraptable}{r}{0.5\textwidth}
    \vspace{-8mm}
    \centering
    \small
    \begin{tabular}{ccccc|c}
        \toprule
         \textbf{\# L1} & \textbf{\# L2} & \textbf{\# L3} & \textbf{\# L4} & \textbf{\# L5} & \textbf{\# Total} \\
        \midrule
        43 & 90 & 105 & 128 & 134 & 500 \\
        \bottomrule
    \end{tabular}
    \vspace{-2mm}
    \caption{\small Data statistics of the MATH test set.}
    \label{tab:math_test}
    \vspace{-8mm}
\end{wraptable}
The distribution of difficulty levels across the 500 test data samples in MATH is listed in Tab.~\ref{tab:math_test}.

\subsection{Training Details}
\label{sec:training_details}
For supervised fine-tuning in Stage I, we adopt LoRA to fine-tune the strong model $\M$ with a learning rate of $1 \times 10^{-4}$ and search for weight decay in the set $\{0, 0.01\}$. We run 2 epochs on GSM8K and 3 epochs on MATH, with a batch size of 8. In Stage II, we employ two preference optimization methods. For DPO, we train the enhanced strong model $\M_\text{plus}$ with a learning rate of $1 \times 10^{-5}$ and run 1 epoch. For ORPO, we search for $\beta$ in the set $\{0.1, 0.5, 1.0\}$ with a learning rate of $3 \times 10^{-5}$ and run 1 epoch. For experiments on OlympicArena using Llama3 family, the hyperparameters are consistent with those used for GSM8K. All experiments are conducted using A100 GPUs. Moreover, when constructing contrastive samples in Stage II, we sample $n=10$ responses at $\text{temperature}=1.0$, and use a confidence threshold of $\tau=0.6$. Normally, we evaluate using greedy decoding. For calculating pass@k, we set $k=10$ and $\text{temperature}=1.0$.

\subsection{Additional Analysis}
\label{sec:appendix_analysis}

\subsubsection{Diversity Analysis}
\label{sec:diversity_analysis}

\begin{figure*}[ht]
    \centering
    \includegraphics[width=0.9\linewidth]{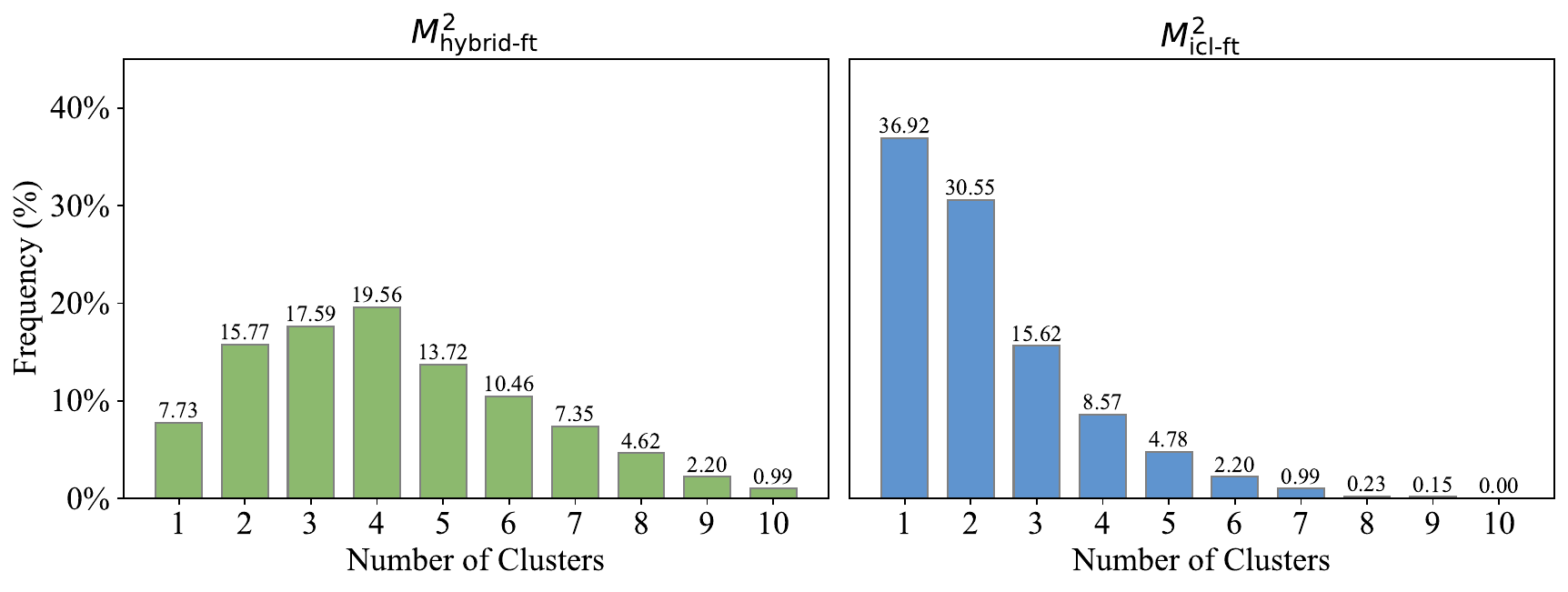}
    \caption{\small Frequency distribution of the number of distinct solutions on GSM8K supervised by Llama2-7b.}
    \label{fig:diversity_distribution}
\end{figure*}

To investigate why $\M_\text{hybrid-ft}$ achieves high pass@k scores despite lower greedy decoding results, we explore the diversity of responses generated by $\M_\text{hybrid-ft}$ and $\M_\text{icl-ft}$. We specifically examine the frequency distribution of the number of distinct solutions for each question across the two strong model checkpoints.

Given a question from $\D_{\text{gold}, 2}$, we sample $n = 10$ responses at $\text{temperature}=1.0$ for each checkpoint. We consider two responses distinct if their ROUGE-L similarity is less than 0.7. We then compute the number of clusters formed by these distinct responses and plot their frequency distribution in Fig.~\ref{fig:diversity_distribution}.

As shown in Fig.~\ref{fig:diversity_distribution}, $\M_\text{icl-ft}^2$ tends to produce nearly the same sampled responses for each question in more than 36\% of the instances. This indicates a limited exploration of problem-solving paths and difficulty in generating diverse, correct solutions during the sampling process. In contrast, $\M_\text{hybrid-ft}^2$ generates a variety of responses, increasing its hit rate with multiple sampling and thus achieving higher pass@k scores. Additionally, diverse solutions are crucial for robust outcomes and model generalization \citep{yu2024flowreasoningefficienttraining,wu2024progressregressselfimprovementreversal}. In Stage II, diverse solutions also ensure the distinction between positive and negative samples, demonstrating the rationale for selecting $\M_\text{hybrid-ft}^2$ for preference optimization in Stage II.

\subsubsection{Training Accuracy of Stage I}
\label{sec:train_acc}

\begin{wraptable}{r}{0.5\textwidth}
    \centering
    \small
    \begin{tabular}{llcc}
        \toprule
         & & \textbf{Final Answer} & \textbf{Process-Level} \\
        \midrule
        \multicolumn{4}{l}{\textbf{GSM8K}} \\
        \midrule
        \multirow{2}{*}{Llama2-7b} & $\hat{\D}_{\text{weak}}^1$ & 89.82 & 72.50 \\
         & $\hat{\D}_{\text{icl}}^1$ & 89.82 & 76.50 \\
        \addlinespace[0.5em]
        \multirow{2}{*}{Gemma-2b} & $\hat{\D}_{\text{weak}}^1$ & 87.97 & 73.10 \\
         & $\hat{\D}_{\text{icl}}^1$ & 87.97 & 73.80 \\
        \addlinespace[0.5em]
        \multirow{2}{*}{Mistral-7b} & $\hat{\D}_{\text{weak}}^1$ & 92.38 & 80.10 \\
         & $\hat{\D}_{\text{icl}}^1$ & 92.38 & 77.90 \\
        \midrule
        \multicolumn{4}{l}{\textbf{MATH}} \\
        \midrule
        \multirow{2}{*}{Llama2-7b} & $\hat{\D}_{\text{weak}}$ & 46.11 & 32.04 \\
         & $\hat{\D}_{\text{icl}}$ & 46.11 & 39.22 \\
        \addlinespace[0.5em]
        \multirow{2}{*}{Gemma-2b} & $\hat{\D}_{\text{weak}}$ & 30.40 & 26.30 \\
         & $\hat{\D}_{\text{icl}}$ & 31.90 & 29.90 \\
        \addlinespace[0.5em]
        \multirow{2}{*}{Mistral-7b} & $\hat{\D}_{\text{weak}}$ & 24.75 & 21.50 \\
         & $\hat{\D}_{\text{icl}}$ & 25.25 & 25.60 \\
        \bottomrule
    \end{tabular}
    \caption{\small Training accuracy of Stage I.}
    \label{tab:train_acc}
\end{wraptable}

Tab.~\ref{tab:train_acc} presents the final answer accuracy and process-level accuracy for both weak data and icl data utilized in the initial round.\footnote{The relatively low accuracy observed in MATH explains why we choose to perform \textit{one} round of iteration.} To compute process-level accuracy, we randomly sample a maximum of 1,000 training sample from each of weak data and icl data, and evaluate them using GPT-4o following \citet{DBLP:journals/corr/abs-2404-05692,DBLP:journals/corr/abs-2312-17080}, the prompt we use is illustrated in Tab.~\ref{tab:prompt}. Accuracy at this level is determined strictly on the basis that there are no errors throughout the intermediate reasoning steps.

From the results we can see that despite having consistent final answer accuracy (with the exceptions of Gemma-2b and Mistral-7b on MATH using augmented training data), there are noticeable differences in process-level performance, leading to variations in the effectiveness of $\M_{\text{weak-ft}}$ and $\M_{\text{icl-ft}}$. Moreover, it is counterintuitive that models trained on icl data with relatively low process-level accuracy achieve higher performance. This might be because the models prefer self-generated solutions and can more effectively learn those that better align with their inherent distribution \citep{DBLP:journals/corr/abs-2404-13076,DBLP:journals/corr/abs-2402-11192,DBLP:journals/corr/abs-2402-12219}.

\begin{table*}[h]
    \centering
    \small
    \begin{tabular}{p{\linewidth}}
        \toprule
        \vspace{-2mm}
        Question:\\
        \{question\}\\
        \\
        Student Solution:\\
        \{solution\}\\
        \\
        Your task involves three parts:\\
        1. **Step-by-step Evaluation:** Go through the student solution carefully and identify key errors and potential misunderstandings that led to the incorrect solution.\\
        2. **Final Judgement:**  Provide an overall judgement on the correctness of the student's solution.\\
        3. **First Error Step:** If the solution is incorrect, generate the step number where the first error occurs, otherwise generate N/A here.\\
        \\
        Here's the format I want:\\
        Step-by-step Evaluation: [Provide a step by step examination of the student solution and identify key errors and misunderstandings here.]\\
        Final Judgement: [Insert only **correct** or **wrong** here]\\
        First Error Step: [Insert either N/A or the step number where the first error occurs]\\
        \\
        Please follow this format without any additional introductory or concluding statements.\\
        \bottomrule
    \end{tabular}
    \caption{\small Prompt used to evaluate process-level accuracy.}
    \label{tab:prompt}
\end{table*}

\clearpage

\subsection{Additional Experiments}

\begin{table}[ht]
    \centering
    \begin{minipage}{0.45\textwidth}
        \vspace{0pt}
        \centering
        \small
        \begin{tabular}{llcc}
            \toprule
             & & \textbf{Greedy Decoding} & \textbf{Pass@k} \\
            \midrule
            \multicolumn{4}{l}{\textbf{GSM8K}} \\
            \midrule
            \multirow{3}{*}{Llama2-7b} & $\M_{\text{weak-ft}}^2$ & 57.47 & 77.26 \\
             & $\M_{\text{icl-ft}}^2$ & \textbf{63.76} & 81.05 \\
             & $\M_{\text{hybrid-ft}}^2$ & 62.62 & \textbf{86.28} \\
            \addlinespace[0.5em]
            \multirow{3}{*}{Gemma-2b} & $\M_{\text{weak-ft}}^2$ & 45.03 & 71.49 \\
             & $\M_{\text{icl-ft}}^2$ & \textbf{60.12} & 80.14 \\
             & $\M_{\text{hybrid-ft}}^2$ & 56.03 & \textbf{85.14} \\
            \addlinespace[0.5em]
            \multirow{3}{*}{Mistral-7b} & $\M_{\text{weak-ft}}^2$ & 66.72 & 85.67 \\
             & $\M_{\text{icl-ft}}^2$ & 66.64 & 84.08 \\
             & $\M_{\text{hybrid-ft}}^2$ & \textbf{68.39} & \textbf{88.70} \\
            \midrule
            \multicolumn{4}{l}{\textbf{MATH}} \\
            \midrule
            \multirow{3}{*}{Llama2-7b} & $\M_{\text{weak-ft}}^1$ & 10.80 & 34.80 \\
             & $\M_{\text{icl-ft}}^1$ & 11.80 & \textbf{35.00} \\
             & $\M_{\text{hybrid-ft}}^1$ & \textbf{14.00} & 33.60 \\
            \addlinespace[0.5em]
            \multirow{3}{*}{Gemma-2b} & $\M_{\text{weak-ft}}^1$ & \textbf{14.80} & 38.80 \\
             & $\M_{\text{icl-ft}}^1$ & 13.60 & 33.60 \\
             & $\M_{\text{hybrid-ft}}^1$ & \textbf{14.80} & \textbf{39.60} \\
            \addlinespace[0.5em]
            \multirow{3}{*}{Mistral-7b} & $\M_{\text{weak-ft}}^1$ & 10.80 & 34.20 \\
             & $\M_{\text{icl-ft}}^1$ & \textbf{15.60} & 31.60 \\
             & $\M_{\text{hybrid-ft}}^1$ & 14.20 & \textbf{38.40} \\
            \bottomrule
        \end{tabular}
        \caption{\small Greedy decoding and pass@k results ($k=10$ and $\text{temperature}=1.0$) for the three variants of enhanced strong models obtained through weak-icl fine-tuning. The best results are in \textbf{bold}.}
        \label{tab:passk}
    \end{minipage}
    \hspace{0.05\textwidth}
    \begin{minipage}{0.45\textwidth}
        \vspace{-3mm}
        \centering
        \small
        \begin{tabular}{lcc}
            \toprule
             & \textbf{Test Acc.} & \textbf{\# Training Data} \\
            \midrule
            \multicolumn{3}{l}{\textbf{Gemma-2b}} \\
            SFT on Full Weak & 10.00 & 6,000 \\
            SFT on Gold Weak & 15.60 & 644 \\
            $\M_{\text{weak-ft}}^1$ & 11.00 & 448 \\
            $\M_{\text{icl-ft}}^1$ & 11.40 & 448 \\
            $\M_{\text{hybrid-ft}}^1$ & 13.20 & $448 \times 2$ \\
            \midrule
            \multicolumn{3}{l}{\textbf{Mistral-7b}} \\
            SFT on Full Weak & 14.40 & 6,000 \\
            SFT on Gold Weak & 16.60 & 861 \\
            $\M_{\text{weak-ft}}^1$ & 12.40 & 584 \\
            $\M_{\text{icl-ft}}^1$ & 15.60 & 584 \\
            $\M_{\text{hybrid-ft}}^1$ & 14.20 & $584 \times 2$ \\
            \bottomrule
        \end{tabular}
        \vspace{-2mm}
        \caption{\small Stage I results on MATH without augmenting training data. ``Test Acc.'' refers to Test Accuracy.}
        \label{tab:stagei_math_results}
        \vspace{4mm}
        \begin{tabular}{lcc}
            \toprule
            \textbf{Weak Model} & \textbf{Full Weak FT} & \textbf{Weak-ICL FT \strut} \\
            \midrule
            \multicolumn{3}{l}{\textbf{GSM8K}} \\
            Llama2-7b & 22.47 & 78.53 \\
            Gemma-2b & 8.27 & 75.71 \\
            Mistral-7b & 14.63 & 71.38 \\
            \midrule
            \multicolumn{3}{l}{\textbf{MATH}} \\
            Llama2-7b & 10.45 & 71.64 \\
            Gemma-2b & \textcolor{red}{-25.81} & 64.52 \\
            Mistral-7b & 19.05 & 28.57 \\
            \bottomrule
        \end{tabular}
        \vspace{-2mm}
        \caption{\small Performance Gap Recovered (PGR) in Stage I.}
        \label{tab:pgr}
    \end{minipage}
\end{table}

\subsubsection{Details of Stage I on MATH}
\label{sec:stagei_math}

In the Stage I experiment conducted on the MATH dataset, it is found that the amount of training data selected via final answer consistency is so limited that the strong model can hardly learn the effective features through supervised fine-tuning. To address this, we randomly sample additional inconsistent data. Based on the weak model's performance (Llama-7b $<$ Gemma-2b $<$ Mistral-7b on MATH), we supplement the data (both $\hat{\D}_\text{weak}$ and $\hat{\D}_\text{icl}$) to 1,000 instances for Gemma-2b and 2,000 instances for Mistral-7b, and present the results in Fig.~\ref{fig:main_results}. The original amount of training data and test accuracy for these two weak models are shown in Tab.~\ref{tab:stagei_math_results}.

\subsubsection{Pass@k Results}
\label{sec:passk}
Tab.~\ref{tab:passk} summarizes the greedy decoding and pass@k results for the three variants of enhanced strong models obtained through weak-icl fine-tuning. Notably, $\M_{\text{hybrid-ft}}$ utilizes a training set that combines those used by $\M_{\text{weak-ft}}$ and $\M_{\text{icl-ft}}$. The results indicate that $\M_{\text{hybrid-ft}}$ outperforms its counterparts in terms of pass@k, achieving superior pass@k scores with margins of up to 5.23 points. The only exception occurs in the MATH dataset supervised by Llama2-7b, where the underperformance is likely due to limited training data.

The superior performance of $\M_{\text{hybrid-ft}}$ can be attributed to the diversity of solutions in its training set (verified in \S\ref{sec:diversity_analysis}), validating our approach of adopting the final iteration of $\M_{\text{hybrid-ft}}$ from Stage I for preference optimization in Stage II. It is important to note that while higher pass@k scores suggest greater potential, the true challenge lies in effectively harnessing this potential, particularly in the weak-to-strong setting where no ground truths are available. Our proposed weak-to-strong preference optimization in Stage II successfully addresses this challenge, transforming theoretical potential into tangible performance gains in greedy decoding, as proved in \S\ref{sec:stageii_results}.

\subsubsection{PGR of Stage I}

\citet{DBLP:journals/corr/abs-2312-09390} propose a new metric called performance gap recovered (PGR) to measure the fraction of the performance gap that can be recovered through weak supervision, as illustrated in Eq.~\ref{eq:1}. Tab.~\ref{tab:pgr} displays the results of the naive full weak fine-tuning (i.e., Full Weak FT) and our best weak-icl fine-tuning (i.e., Weak-ICL FT) in terms of PGR, which also demonstrate that our method can outperform the simple competitor. However, the variations in PGR across different weak models do not provide meaningful insights. In the experiments described in the main text, we use test accuracy instead to provide a more detailed depiction of model performance.
\begin{align}
\text{PGR} = \frac{\text{weak-to-strong} - \text{weak floor}}{\text{strong ceiling} - \text{weak floor}}.
\label{eq:1}
\end{align}

\subsubsection{Effect of SFT Data}

\begin{wraptable}{r}{0.5\textwidth}
    \vspace{-8mm}
    \centering
    \small
    \begin{tabular}{cll}
        \toprule
        \textbf{Weak Model} & \textbf{SFT Data} & \textbf{Test Accuracy} \\
        \midrule
        \multirow{6}{*}{Llama2-7b} & Full Weak & 42.38 \\
         & Gold Weak & 54.21 (\textcolor{GreenCheck}{+11.83}) \\
         & Our Weak & 53.68 (\textcolor{GreenCheck}{+11.30}) \\
         \cmidrule(lr){2-3}
         & Full ICL & 59.14 \\
         & Gold ICL & 64.29 (\textcolor{GreenCheck}{+5.15}) \\
         & Our ICL & 61.71 (\textcolor{GreenCheck}{+2.57}) \\
        \midrule\midrule
        \multirow{6}{*}{Gemma-2b} & Full Weak & 29.04 \\
         & Gold Weak & 46.40 (\textcolor{GreenCheck}{+17.36}) \\
         & Our Weak & 42.91 (\textcolor{GreenCheck}{+13.87}) \\
         \cmidrule(lr){2-3}
         & Full ICL & 58.61 \\
         & Gold ICL & 63.86 (\textcolor{GreenCheck}{+5.25}) \\
         & Our ICL & 59.21 (\textcolor{GreenCheck}{+0.60}) \\
        \midrule\midrule
        \multirow{6}{*}{Mistral-7b} & Full Weak & 61.33 \\
         & Gold Weak & 67.55 (\textcolor{GreenCheck}{+6.22}) \\
         & Our Weak & 65.96 (\textcolor{GreenCheck}{+4.63}) \\
         \cmidrule(lr){2-3}
         & Full ICL & 62.32 \\
         & Gold ICL & 66.64 (\textcolor{GreenCheck}{+4.32}) \\
         & Our ICL & 65.43 (\textcolor{GreenCheck}{+3.11}) \\
        \bottomrule
    \end{tabular}
    \caption{\small Detailed results of Stage I on GSM8K.}
    \label{tab:sft_data_results}
    \vspace{-10mm}
\end{wraptable}

Tab.~\ref{tab:sft_data_results} presents more detailed comparative experimental results of Stage I on GSM8K. ``Full Weak'' denotes full weak fine-tuning, ``Our Weak'' is equivalent to $\M_{\text{weak-ft}}^1$, and ``Our ICL'' is equivalent to $\M_{\text{icl-ft}}^1$. ``Gold Weak'' refers to the scenario where weak data with correct final answers are filtered and used for supervised fine-tuning, which is impossible in the weak-to-strong setting and just used for experimental analysis. Similarly, ``Gold ICL'' refers to the scenario where solutions with correct final answers, generated by the strong model via weak ICL, are filtered.

Compared to using a large volume of noisy data (i.e., Full Weak and Full ICL), reducing the data quantity while enhancing data quality can significantly improve the accuracy of the trained model, with potential gains over 17 points. Although our method performs slightly lower than the gold results, it proves highly effective and stable in scenarios where obtaining the ground truth is impossible.

\vspace{8mm}
\subsubsection{Zero-Shot Results}
\label{sec:zero-shot}

\begin{wraptable}{r}{0.5\textwidth}
    \vspace{-3mm}
    \centering
    \small
    % \resizebox{\columnwidth}{!}{
    \begin{tabular}{l@{\hspace{50pt}}c}
        \toprule
         & \textbf{Test Accuracy} \\
        \midrule
        \multicolumn{2}{l}{\textbf{GSM8K}} \\
        \midrule
        Llama2-70b wo CoT & 12.36 \\
        Llama2-70b w/ CoT & 18.35 \\
        \midrule
        \multicolumn{2}{l}{\textbf{MATH}} \\
        \midrule
        Llama2-70b wo CoT & 6.40 \\
        Llama2-70b w/ CoT & 7.20 \\
        \bottomrule
    \end{tabular}
    \caption{\small Zero-Shot Results of Llama2-70b on GSM8K and MATH.}
    \label{tab:zero-shot}
\end{wraptable}

To obtain zero-shot performance, we follow \citet{DBLP:conf/nips/KojimaGRMI22} using a two-stage prompting approach. Specifically, we use the first prompt to extract a full reasoning path, where ``wo CoT'' denotes the standard prompt ``Question: \{question\}\textbackslash nAnswer:'', while ``w/ CoT'' denotes the CoT prompt ``Question: \{question\}\textbackslash nLet's think step by step.\textbackslash nAnswer:''. Then we use the second prompt, which concatenates ``The answer is'' with the generated reasoning path, to extract the answer in the correct format. The zero-shot results of Llama2-70b on the two reasoning datasets are presented in Tab.~\ref{tab:zero-shot}. We can observe that these results are significantly lower than those achieved with weak ICL. This notably poor zero-shot performance aligns with our hypothesis about the strong model: before any fine-tuning with weak supervision, the strong model's capabilities have not been fully realized.

\end{document}